\documentclass[letterpaper]{article} % DO NOT CHANGE THIS
\usepackage{sty/aaai2026}  % DO NOT CHANGE THIS
\usepackage{times}  % DO NOT CHANGE THIS
\usepackage{helvet}  % DO NOT CHANGE THIS
\usepackage{courier}  % DO NOT CHANGE THIS
\usepackage[hyphens]{url}  % DO NOT CHANGE THIS
\usepackage{graphicx} % DO NOT CHANGE THIS
\usepackage{natbib}  % DO NOT CHANGE THIS AND DO NOT ADD ANY OPTIONS TO IT
\usepackage{caption} % DO NOT CHANGE THIS AND DO NOT ADD ANY OPTIONS TO IT
\usepackage{multirow}

\usepackage[T1]{fontenc}
\usepackage{ifthen}
\usepackage{algorithm}
\usepackage[noEnd=true,indLines=true,commentColor=black]{sty/algpseudocodex}
\usepackage{amsmath,amssymb,amsthm}
\usepackage{booktabs}
\usepackage{subcaption}
\usepackage{cleveref}
\usepackage{siunitx}
\usepackage{makecell}
\usepackage{sty/custom}

\nocopyright
\usepackage{newfloat}
\usepackage{listings}
\DeclareCaptionStyle{ruled}{labelfont=normalfont,labelsep=colon,strut=off} % DO NOT CHANGE THIS
\floatstyle{ruled}
\newfloat{listing}{tb}{lst}{}
\floatname{listing}{Listing}
\title{Alternating Target–Path Planning for Scalable Multi-Agent Coordination}
\author{Yu Kumagai$^{1,2}$, Keisuke Okumura$^{2}$}
\affiliations{
  $^1$Hitotsubashi University, Japan
  \\
  $^2$National Institute of Advanced Industrial Science and Technology (AIST), Japan
  \\
  \texttt{\{yu.kumagai,okumura.k\}@aist.go.jp}
}

\begin{document}

\maketitle

\begin{abstract}
The concurrent \emph{target assignment and pathfinding (TAPF)} problem extends multi-agent pathfinding (MAPF) by asking planners to allocate distinct targets and collision-free paths to agents.
Prior work on TAPF has relied exclusively on Conflict-Based Search (CBS), which tightly couples target assignment and pathfinding, resulting in compute-intensive,  non-scalable solutions.
In contrast, we propose an iterative refinement framework that decouples target assignment from pathfinding.
Our framework builds on modern, fast, suboptimal MAPF solvers, such as LaCAM.
Specifically, within a given time budget, it repeatedly solves MAPF for the current target assignment, identifies bottleneck agents via MAPF feedback, and refines the assignment.
Empirical results show that feedback-driven reassignment loop is effective, enabling our framework to scale well beyond the reach of the state-of-the-art CBS-based solver while maintaining decent solution quality.
This represents a solid step toward practical, large-scale TAPF suitable for real-world setups.
\end{abstract}

\section{Introduction}

The concurrent \emph{target assignment and pathfinding (TAPF)} problem~\cite{CBM, Barták_Ivanová_Švancara_2021} for multiple agents on a graph is an appealing extension of the classical multi-agent pathfinding (MAPF) formulation~\cite{stern2019def}.
In TAPF, each agent is given a set of possible target locations, and a planner must not only generate collision-free paths but also assign distinct targets to each agent.
This formulation is considered more reflective of real-world logistics scenarios than vanilla MAPF, as many such domains are \emph{partially agent-agnostic}, meaning the specific agent-target pairing is to some extent flexible, as long as all targets are completed.

The target assignment, a subproblem of TAPF, is a well-established field with classical solutions such as the Hungarian algorithm~\cite{Hungarian}, bottleneck matching~\cite{gross1959bottleneck}, and auction-based approaches~\cite{auction-based}.
MAPF, on the other hand, has seen significant developments in the last decade.
With the progress in MAPF research, several TAPF algorithms have been proposed~\cite{CBM, (e)cbs-ta, ita-cbs, ITA-ECBS}.
However, existing TAPF approaches are exclusively built upon Conflict-Based Search (CBS)~\cite{SHARON201540} and its variants, which guarantee optimal or bounded-suboptimal solutions but suffer from limited scalability.
As a result, current TAPF solutions struggle to handle large instances involving hundreds of agents within practical time constraints.
In contrast, recent developments in unbounded suboptimal MAPF algorithms---such as PIBT~\cite{okumura2022priority}, LNS2~\cite{LiAAAI22}, and LaCAM~\cite{okumura2023lacam}---have demonstrated remarkable scalability.
These methods prioritize computational efficiency over guaranteed optimality, enabling them to solve large instances within a realistic timeframe.
This scalability gap motivates a different approach to TAPF.
{
  \begin{figure}[t!]
    \centering
    \includegraphics[width=1\linewidth]{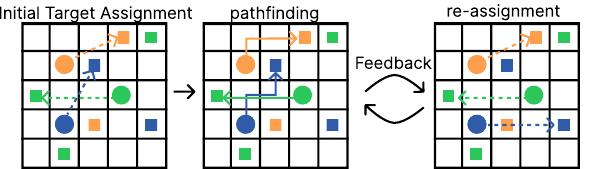}
    \caption{Iterative refinement framework for TAPF.}
    \label{fig:framework}
  \end{figure}
}

To this end, we propose an \emph{iterative refinement framework} that decouples target assignment from pathfinding, leveraging modern, fast, suboptimal MAPF solvers as subroutines.
As \cref{fig:framework} illustrates, within a given time budget, our framework repeatedly solves MAPF for the current assignment, analyzes the resulting paths to identify bottleneck agents, and uses this feedback to refine the assignment.

While this approach is conceptually straightforward, key design choices remain in \emph{(i)~the feedback mechanism}, which determines which target assignments should be reconsidered based on the previous pathfinding, and \emph{(ii)~the reassignment strategy}, which determines how to reallocate targets based on the feedback.
For each perspective, we propose two reasonable methods.

Specifically, for \emph{(i)}~the feedback mechanism, we present \emph{Delay-Based Selection (DBS)}, which identifies bottleneck agents based on their delays from ideal paths, and \emph{Spectral Bottleneck Sampling (SBS)}, which leverages spectral analysis to identify groups of spatially correlated agents experiencing collective delays.
For \emph{(ii)}~the reassignment strategy, we present \emph{PIBT}, which adapts PIBT~\cite{okumura2022priority} to target reassignment leveraging its efficient collision-free allocation mechanism, and \emph{Local Hungarian}, which applies the Hungarian algorithm to a small subgroup of agents to achieve locally optimal reassignment.

We evaluate these combinations on the commonly-used MAPF benchmark~\cite{stern2019def} and demonstrate that our framework---especially the combination of DBS and Local Hungarian---scales well beyond the reach of the state-of-the-art CBS-based TAPF solver~\cite{ITA-ECBS} while maintaining competitive solution quality.
This moves TAPF meaningfully closer to practical, scalable deployment in real-world applications.
The code is available in \url{https://github.com/Ukuma012/ir-tapf}.

\section{Preliminaries}
\label{sec:preliminaries}

\subsection{Problem Definition}
We consider a system consisting of a graph $G = (V, E)$ and a set of agents $A = \{1, \ldots, n\}$.
Let $\mathcal{T}$ be a binary assignment matrix of size $n \times |V|$, where $\mathcal{T}[i][j] = 1$ means that a vertex $v \in V$ of index $j$ can be the target of an agent $i \in A$. For convenience, we will use the notation $\mathcal{T}[i][v]$.

A \emph{feasible assignment} is an injective function, i.e., matching $\mathcal{M}: A \mapsto V$ that satisfies $\mathcal{T}[i][\mathcal{M}(i)] = 1$ for each agent $i$. A \emph{TAPF instance} is defined by a graph ${G}$, agents $A$, starts $s_i \in V$ for each agent $i \in A$, and a target assignment matrix $\mathcal{T}$ that has at least one feasible assignment.
A \emph{solution} to TAPF is a feasible assignment $\mathcal{M}$ and a collection of collision-free paths $\Pi = \{\pi_1, \ldots, \pi_n\}$ where $\pi_i$ is the path for agent $i$.
For each agent $i$, its path must begin at $s_i$ and end at $\mathcal{M}(i)$.
Agent movement models and collisions follow the classical MAPF formulation~\cite{stern2019def}: i.e., at each time step, an agent can either move to an adjacent vertex or remain at its current vertex.
Two types of conflicts must be avoided:
\emph{(i)}~\emph{vertex conflict}, where two agents occupy the same vertex simultaneously, and \emph{(ii)}~\emph{edge conflict}, where two agents traverse the same edge in opposite directions.

In this formulation, the solution cost is assessed by \emph{flowtime} (aka. sum-of-costs), which sums the travel time of each agent until it stops at the target location and no longer moves.
Within limited planning time, we are ultimately interested in finding flowtime-minimizing solutions.

We use $\dist(u, v)$ to denote the shortest path distance between vertices $u$ and $v$ on graph $G$.
For a path $\pi = \langle v_0, v_1, \ldots, v_T \rangle$, we define $\cost(\pi) = T$ as the path cost (i.e., the travel time).

\subsection{Related Work}

\paragraph{MAPF.}
While early work on MAPF focused on developing efficient optimal algorithms such as CBS~\cite{SHARON201540}, recent research has increasingly shifted toward suboptimal yet highly scalable, real-time methods.
Among these suboptimal approaches, \emph{LaCAM (lazy constraints addition search)}~\cite{okumura2023lacam,okumura2024lacam3} has gained notable attention due to its reliability in delivering solutions quickly even for large numbers of agents operating in densely populated environments.
These ultra-fast MAPF methods have opened up novel multi-robot research directions, such as advanced planning-execution frameworks~\cite{zhang2024planning} and MAPF-based multi-robot control in continuous spaces~\cite{shankar2025lf,okumura2026concrete}.
Our TAPF study also aligns with this direction, leveraging fast MAPF solvers as subroutines to address challenging multi-agent planning problems.

\paragraph{Unlabeled MAPF.}
While standard MAPF assigns each agent a fixed target, \emph{unlabeled} (or \emph{anonymous}) \emph{MAPF} treats agents as fully interchangeable: any agent can be assigned to any target.
Makespan-optimal unlabeled MAPF is solvable in polynomial time via a flow-network reduction~\cite{yu2013multi}, which jointly optimizes target assignment and pathfinding.
However, this approach lacks generality and cannot be applied to TAPF.
By contrast, our approach is more closely aligned with suboptimal methods for unlabeled MAPF~\cite{tswap}, which decouple initial target assignment from pathfinding to achieve better real-time planning performance.

\paragraph{TAPF.}
This setting generalizes both labeled and unlabeled MAPF.
It can be viewed as unlabeled MAPF with agent-specific target sets, where the extreme case is labeled MAPF, in which each agent has exactly one designated target.
Several TAPF algorithms have been proposed, primarily built upon CBS.
For example, CBM~\cite{CBM}, CBS-TA~\cite{(e)cbs-ta}, and ITA-CBS~\cite{ita-cbs} provide optimal solutions.
Their bounded suboptimal versions~\cite{(e)cbs-ta,ITA-ECBS} have also been proposed, dropping optimality for faster computation.
While these CBS-based approaches provide solution quality guarantees, this guarantee comes at the cost of limited scalability.
As shown in our experiments (\cref{sec:evaluation}), the state-of-the-art bounded suboptimal algorithm ITA-ECBS~\cite{ITA-ECBS} is unable to solve instances beyond 250 agents within a reasonable time.
This motivates us to develop practical solutions.

\section{Iterative Refinement Framework}
\label{sec:algorithm}
Our approach to TAPF is straightforward.
The framework begins with an (arbitrary) initial target assignment, rapidly solves the corresponding MAPF problem, analyzes the paths to identify problematic assignments as feedback, reassigns targets, and repeats this process until reaching a given time limit.
This iterative style has been underexplored, as prior work has focused on tightly coupled target assignment and pathfinding within unified CBS-based framework.
Meanwhile, the emergence of ultra-fast MAPF algorithms enables us to quickly estimate the quality of a given target assignment.
Using this feedback, we aim to gradually refine the target assignment under a tight time constraint.

This framework, which decouples target assignment and pathfinding, not only enjoys the scalability advantage stemming from modern MAPF solvers, but also offers an engineering benefit of \emph{modularity}.
Indeed, the framework accommodates arbitrary target assignment strategies and MAPF solvers, so future advances in either component can be directly integrated without modifying the other.

\Cref{algo:framework} formalizes the framework, detailed below.

\paragraph{Initial Target Assignment.}
The process at \cref{algo:framework:initial-assignment} could be arbitrary, but it is desirable to be computationally efficient while assigning each agent to a nearby target, as this provides a strong starting point for subsequent refinement.
Although the polynomial-time, optimal Hungarian algorithm exists~\cite{Hungarian}, its $O(n^3)$ time complexity is impractical for large instances.
To this end, we employ a greedy assignment strategy adopted in a scalable unlabeled-MAPF algorithm~\cite{tswap}.
This method sorts all feasible agent-target pairs by the shortest-path distance and greedily assigns them in order.
It then iteratively improves the assignment through pairwise swaps: if swapping targets between agents $i$ and $j$ reduces the sum of the path lengths and is feasible, the swap is performed.
This process repeats until no beneficial swap remains.

\paragraph{Target Refinement Iteration.}
Each iteration consists of two steps.
First, a \emph{feedback mechanism} analyzes the paths from the previous MAPF solution to identify an agent for reassignment (\cref{algo:framework:identify}).
Next, based on the identified agent, the framework reassigns targets to a subset of the entire team (\cref{algo:framework:reasign}).
The quality of the new assignment is then evaluated by solving its corresponding MAPF problem (\cref{algo:framework:mapf}).
Throughout this process, the framework tracks the best overall solution found so far (\cref{algo:framework:track}).
The framework terminates when interrupted (e.g., by a time limit), returning the best solution found.

\paragraph{Final Path Optimization.}
After the target refinement terminates, the framework can optionally perform a final path optimization (\cref{algo:framework:path-opt}).
While the target refinement benefits from lightweight MAPF solvers for computational efficiency, the resulting paths are typically suboptimal.
Therefore, applying a computationally heavier solver to obtain the final paths can further improve the overall solution quality.
In our implementation, vanilla LaCAM is used during the target-refinement stage, while its engineered variant, LaCAM3~\cite{okumura2024lacam3}, which provides better solution quality with a slight increase in computational overhead, is employed in the final step.

\paragraph{Multi-Bottleneck Evaluation.}
While \cref{algo:framework} presents a naive version that selects a single bottleneck agent per iteration (Lines~\ref{algo:framework:identify}--\ref{algo:framework:mapf}; blue-colored), we can enhance the framework by evaluating multiple candidate reassignments for each iteration.
Specifically, \cref{algo:multiple_bottleneck} provides a drop-in replacement for the corresponding lines in \cref{algo:framework}.
Suppose that the feedback mechanism returns multiple agents (\cref{algo:multiple_bottleneck:identify} in \cref{algo:multiple_bottleneck}).
The framework then generates an alternative assignment for each.
Among these evaluated candidates, we stochastically select one (uniformly at random in the implementation) to continue the refinement.
Rather than always choosing the best candidate, this non-determinism avoids search stagnation and encourages broader exploration of the solution space, ultimately yielding better final solutions.
In practice, this multi-bottleneck evaluation can be parallelized and therefore does not incur significant time overhead compared to the single-bottleneck version.

{
\renewcommand{\cost}{\m{c}}
\renewcommand{\hl}[1]{\textcolor{blue}{#1}}
\begin{algorithm}[t!]
\caption{Iterative Refinement Framework for TAPF}
\label{algo:framework}
\begin{algorithmic}[1]
\small
\Input{TAPF instance $I$}
\Output{task assignment $\assignment$ and collision-free paths $\solutions$}
\State $\assignment \leftarrow \funcname{InitialAssignment}(I)$
\label{algo:framework:initial-assignment}
\State $\solutions \leftarrow \funcname{SolveMAPF}(I, \assignment)$
\State $\assignment\sub{best} \leftarrow \assignment$, $\cost\sub{best} \leftarrow
\funcname{GetCost}(\solutions)$, $\solutions\sub{best} \leftarrow \solutions$

\While{not interrupted}
\State \hl{$a \leftarrow \funcname{GetBottleneckAgent}(I, \assignment, \solutions)$}
\label{algo:framework:identify}
\State \hl{$\assignment \leftarrow \funcname{ReassignTargets}(a, \assignment)$}
\label{algo:framework:reasign}
\State \hl{$\solutions \leftarrow \funcname{SolveMAPF}(I, \assignment)$}
\label{algo:framework:mapf}
\State $c \leftarrow \funcname{GetCost}(\solutions)$
\IfSingle{$c < \cost\sub{best}$}
{
$\assignment\sub{best} \leftarrow \assignment$,
$\cost\sub{best} \leftarrow c$,
$\solutions\sub{best} \leftarrow \solutions$
\label{algo:framework:track}
}
\EndWhile
\State $\solutions\sub{best} \leftarrow \funcname{SolveNearOptimalMAPF}(I, \assignment\sub{best})$
\Comment{optional}
\label{algo:framework:path-opt}
\State \Return $\assignment\sub{best}$, $\solutions\sub{best}$
% %
\end{algorithmic}
\end{algorithm}
}

{
\newcommand{\candidates}{\m{\mathrm{candidates}}}
\begin{algorithm}[t!]
\caption{Multi-Bottleneck Evaluation}
\label{algo:multiple_bottleneck}
\begin{algorithmic}[1]
\Statex {\footnotesize\emph{(replacement for \textcolor{blue}{Lines~\ref{algo:framework:identify}--\ref{algo:framework:mapf}} of \cref{algo:framework}})}
\State $\candidates \leftarrow \emptyset$
\State $B \leftarrow \funcname{GetBottleneckAgents}(I, \assignment, \solutions)$
\label{algo:multiple_bottleneck:identify}
\For{$a \in B$}\Comment{parallelizable}
\State $\assignment' \leftarrow \funcname{ReassignTargets}(a, \assignment)$
\State $\solutions' \leftarrow \funcname{SolveMAPF}(I, \assignment')$
\State $\candidates \gets \candidates \cup \{\langle\assignment', \solutions'\rangle\}$
\EndFor
\State $\assignment, \solutions \leftarrow \funcname{SelectOneRandomly}(\candidates)$
\end{algorithmic}
\end{algorithm}
}

\section{Target Reassignment}
The effectiveness of this framework depends on the method for reassigning targets (\cref{algo:framework:reasign}) and the design of the feedback mechanism (\cref{algo:framework:identify}).
We now describe two approaches for the former: \emph{PIBT} and \emph{Local Hungarian}.

\subsection{PIBT}
When reassigning a bottleneck agent $a\sub{bottleneck}$, a natural difficulty is that its preferred target may already be held by another agent.
To handle such conflicts, we adapt the mechanism of Priority Inheritance with Backtracking (PIBT)~\cite{okumura2022priority}.
In its original pathfinding setting, PIBT resolves positional conflicts by having a higher-priority agent recursively displace lower-priority agents occupying desired vertices; displaced agents, in turn, seek alternative vertices through the same recursive process.
We apply an analogous mechanism to target reassignment: when the $a\sub{bottleneck}$ prefers a target $g$ that is already assigned to another agent, we recursively request it to switch to an alternative target, triggering a chain of reassignments until every displaced agent finds a feasible target or the process fails.

\Cref{algo:pibt_style_displacement} embodies this idea.
Starting from the bottleneck agent, we enumerate all feasible targets in ascending order of distance from its start position, excluding its current target (\cref{algo:pibt_style_displacement:candidates}).
For each candidate target $g$, if $g$ is unassigned or the agent currently holding $g$ can be successfully displaced via a recursive call (\cref{algo:pibt_style_displacement:recurse}), the agent claims $g$.

Crucially, if the recursive call fails, the algorithm does not abort but instead backtracks and proceeds to the next candidate in the sorted list. This backtracking mechanism explores a richer set of feasible reassignments compared to a greedy single-candidate approach.

\paragraph{Time Complexity.}
The time complexity closely follows that of the original PIBT.
Let $K$ denote the maximum number of feasible targets per agent and $F$ the time to evaluate a candidate targets.
The worst-case complexity is $O(|A| \cdot K \cdot F)$, as the displacement may involve up to $|A|$ agents.
In practice, it terminates after far fewer than $|A|$ displacements.

\subsection{Local Hungarian Assignment}
Where targets are densely clustered, the reassignment of one agent significantly affects the optimal choices for others.
To handle such cases, we propose \emph{Local Hungarian} assignment, which jointly optimizes a small subgroup of agents rather than reassigning them individually.
Specifically, given a set of agents identified by the feedback mechanism, we treat this entire set as a subgroup.
For this subgroup, we collect a pool of candidate targets consisting of their current targets and unassigned targets that are not currently assigned to any agent outside subgroup.
We then construct a cost matrix where entry $(i, v)$ represents the shortest path distance from agent $i$'s start position $s_i$ to the candidate target $v \in V$.
The Hungarian algorithm~\cite{Hungarian} is applied to this cost matrix to find the minimum linear-sum cost assignment of the subgroup, while ignoring inter-agent collisions.

Instead of re-optimizing the assignment for all agents, this local optimization strategy focuses the optimization effort on the most critical subset of the entire team.
Consequently, it resolves the identified bottleneck assignment effectively.

\section{Feedback Mechanism}
The second key component is the feedback mechanism.
A straightforward approach would randomly select an agent for reassignment, but it would ignore the valuable information from the pathfinding results.
Instead, we should systematically identify the agent that struggled most, as this agent is likely the cost bottleneck.
We propose two feedback mechanisms, called \emph{Delay-Based Selection (DBS)} and \emph{Spectral Bottleneck Sampling (SBS)}.

\subsection{Delay-Based Selection (DBS)}
\label{sec:dbs}
DBS, representing the simplest yet effective idea, identifies an agent to be reassigned by ranking all agents according to their individual \emph{delay}---how much their actual path cost exceeds the ideal shortest-path cost.
Specifically, for each agent $i$, DBS computes the delay as $\delay_i = \cost(\pi_i) - \dist(s_i, \mathcal{M}(i))$.
The higher $\delay_i$ means that $i$'s path has greater room for improvement.
Such a delay-based score is commonly used in large neighborhood search refinement for MAPF solutions~\cite{li2021anytime,okumura2021iterative}; here we employ it for target reassignment.

Based on this delay measure, DBS selects the top-$m$ agents ($m \ll n$; set to 10 in our experiments), and then randomly chooses $k$ of them as the current bottlenecks.
With the Local Hungarian assignment, we observed that applying these $k$ agents directly within a single reassignment optimization yields better performance; thus, the multi-bottleneck mechanism is disabled with Local Hungarian.

\subsection{Spectral Bottleneck Sampling (SBS)}
\label{sec:sbs}
As an advanced method, we propose SBS, which leverages spectral decomposition to identify groups of spatially correlated agents that exhibit collective delays.
Specifically, it identifies bottleneck agents through two stages:
\emph{(i)}~constructing a \emph{weighted performance discrepancy matrix} $\discrepancy$ that captures both spatial correlation and performance degradation among agent pairs, and
\emph{(ii)}~applying \emph{spectral decomposition} to reveal groups of agents experiencing correlated bottlenecks.
Spectral decomposition is effective because eigenvectors associated with large eigenvalues tend to highlight clusters of agents that mutually interfere in the same congested regions, enabling us to identify more structural bottlenecks.
\Cref{algo:sbs} shows its pseudocode.

{
  \begin{algorithm}[t!]
  \caption{PIBT-style Displacement}
  \label{algo:pibt_style_displacement}
  \begin{algorithmic}[1]
  \small
  \Input{bottleneck agent $a\sub{bottleneck}$, assignment $\assignment$}
  \Output{new assignment $\assignment$}

  \State \Return \Call{PIBT}{$a\sub{bottleneck}$}

  \Function{PIBT}{$a$}
  \State $C \gets \bigl[v \mid \mathcal{T}[a][v] = 1, v \neq \assignment(a)\bigr]$ sorted by $\dist(s_a, v)$
  \label{algo:pibt_style_displacement:candidates}
  \For{$g \in C$}
  \State $a' \gets b \in A$ s.t.\ $\assignment(b) = g$
  \If{$\not\exists\, a'$ \textbf{or} \Call{PIBT}{$a'$} $\neq$ \failure}
  \label{algo:pibt_style_displacement:recurse}
  \State $\assignment(a) \gets g$; \Return $\assignment$
  \EndIf
  \EndFor

  \State \Return \failure
  \EndFunction
  \end{algorithmic}
  \end{algorithm}
}

  {
    \begin{algorithm}[t!]
      \caption{Spectral Bottleneck Sampling (SBS)}
      \label{algo:sbs}
      \begin{algorithmic}[1]
      \small      
        \Input{Instance $I$, assignment $\assignment$, solutions $\solutions$}
        \Output{bottleneck agent $a\sub{bottleneck}$}
        \State $\mymatrix \leftarrow \funcname{ComputePotentialConflictMatrix}(I, \assignment)$
        \label{algo:sbs:potential-conflict-matrix}
        \State $\discrepancy \leftarrow \text{zero matrix of size } n \times
  n$
        \For{$i \leftarrow 1, \ldots, n;\; j \leftarrow i+1, \ldots, n$}
        \State $\discrepancy[i][j], \discrepancy[j][i] \leftarrow \mymatrix[i][j] \times (\delay_i + \delay_j)$
        \label{algo:sbs:performance-discrepancy-matrix}
        \EndFor

        \State $\lambda, V \leftarrow \funcname{SpectralDecomposition}(\discrepancy)$
        \label{algo:sbs:spectral-decomposition}

        \State $a\sub{bottleneck} \leftarrow \argmax_{i} |V[1][i]|$

        \State \Return $a\sub{bottleneck}$
      \end{algorithmic}
    \end{algorithm}
  }

{
  \begin{figure*}[th!]
    \centering
    \includegraphics[width=1\linewidth]{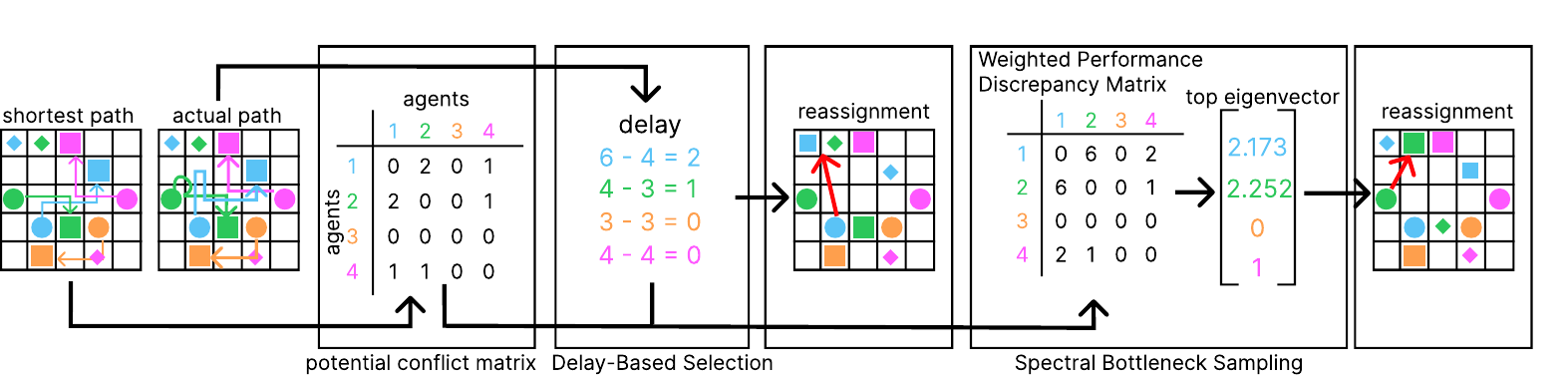}
    \caption{Example of DBS and SBS.
      Circles represent agents, and large squares represent their currently assigned targets. 
      On the left grids, the top left small diamonds show the other assignable targets for agents 1 and 2.}
    \label{fig:example}
  \end{figure*}
}

\paragraph{Weighted Performance Discrepancy Matrix,} $\discrepancy$, aims to capture the group interaction among agents, constructed with two steps (\cref{fig:example}):
\emph{(i)}~A \emph{potential conflict matrix} $\mymatrix$ is computed, where $\mymatrix[i][j]$ counts the number of vertex and edge conflicts between agents $i$ and $j$'s ideal shortest paths (\cref{algo:sbs:potential-conflict-matrix}).
This quantifies spatial correlation: high $\mymatrix[i][j]$ indicates spatiotemporal overlaps between agents $i$ and $j$.
Indeed, such a score is often employed in recent MAPF methods to mitigate congestion~\cite{zhang2024guidance,Kato_2025,arita2025local};
\emph{(ii)}~We next construct the matrix \discrepancy by combining spatial correlation with performance degradation:
$\discrepancy[i][j] = \mymatrix[i][j] \times (\delay_i + \delay_j)$, where $\delay_i$ is delay as defined in DBS.
From its construction, $\discrepancy[i][j]$ is high when $i$ and $j$ are both spatially correlated \emph{and} experiencing delays.
Note that the matrix symmetry (i.e., $\discrepancy[i][j] = \discrepancy[j][i]$) makes it suitable for spectral analysis.

\paragraph{Spectral Decomposition.}
We next apply spectral decomposition to the obtained matrix $\discrepancy$ (\cref{algo:sbs:spectral-decomposition}), resulting in
$\discrepancy = V \Lambda V^T$
where $\Lambda$ contains eigenvalues $\lambda_1 \geq \lambda_2 \geq \ldots \geq \lambda_n$, and $V = [v_1, v_2, \ldots, v_n]$ contains corresponding eigenvectors.
The eigenvalue $\lambda_k$ represents the strength of the $k$-th principal mode of correlated discrepancy, while the eigenvector $v_k$ identifies which agents participate in this mode.
Agents with high absolute values in $v_k$ form a group, capturing those that exhibit correlated delays.

\paragraph{Bottleneck Selection.}
For the single-bottleneck selection, we choose the agent with the largest absolute value in the first eigenvector $V[1]$, which corresponds to the largest eigenvalue $\lambda_1$.
For the multi-bottleneck case, we consider two styles:
\emph{(i)}~selecting $k$ eigenvector groups from the top-ranked groups and picking one agent with the highest absolute eigenvector value from each group; or
\emph{(ii)}~using the single highest eigenvector and selecting the top-$k$ agents with the highest absolute eigenvector values within that group.
Empirically, the former works well with PIBT, whereas the latter is more effective with Local Hungarian.

\paragraph{Time Complexity.}
SBS has $O(n^2)$ overhead for constructing the matrix \discrepancy (Lines \ref{algo:sbs:potential-conflict-matrix}\nobreakdash--\ref{algo:sbs:performance-discrepancy-matrix}) and $O(n^3)$ for eigenvalue decomposition (\cref{algo:sbs:spectral-decomposition}).
Meanwhile, \discrepancy exhibits high sparsity (90-98\% in our experiments) due to two factors: \emph{(i)}~spatially distant agents have zero potential conflicts ($\mymatrix[i][j] = 0)$, and \emph{(ii)}~agents that take ideal path contribute zero performance delays ($\delay_i = 0)$.
Therefore, we can use iterative sparse eigenvalue solvers, such as the Lanczos method~\cite{Lanczos:1950zz}.
For finding the largest eigenvalue and its corresponding eigenvector, this reduces complexity from $O(n^3)$ to $O(zm)$ where $z \ll n^2$ is the number of non-zero elements and $m$ is the hyperparameter iteration count.
Implementation-wise, we further employ sub-sampling of the entire team, reducing the matrix size from $n \times n$ to $s \times s$ where $s \ll n$ ($s=100$ in the experiments).
The sub-sampling procedure first collects agents whose start positions are close to the bottleneck agents identified in the previous round, based on the Manhattan distance, and then randomly selects the remaining agents.
This approach preserves scalability while still providing sufficiently informative feedback for the assignment process.

{
\newcommand{\yA}{-0.16\linewidth}
\newcommand{\yB}{0.082\linewidth}
\newcommand{\xA}{0.15\linewidth}
\newcommand{\xB}{0.16\linewidth}
\newcommand{\xC}{0.01\linewidth}

\newcommand{\entry}[5]{
\begin{scope}[shift={(#4 * 0.33 * \linewidth,- #5 * 0.38 * \linewidth)}]
%
% hotspot
\node[anchor=north west](hotspot) at (0, 0)
{\includegraphics[width=\xA]{fig/raw/hotspot/comparison_#1}};
\node[anchor=north west] at (0, \yA)
{\includegraphics[width=\xA]{fig/raw/hotspot/imprv_#1}};
\node[] at ($(hotspot) + (\xC, \yB)$) {\scenHotspot};
%
% random
\node[anchor=north west](random) at (\xA, 0)
{\includegraphics[width=\xA]{fig/raw/random/comparison_#1}};
\node[anchor=north west] at (0\xA, \yA)
{\includegraphics[width=\xA]{fig/raw/random/imprv_#1}};
\node[] at ($(random) + (\xC, \yB)$) {\scenRandom};
%
% labels
\node[] at (\xB, -0.153\linewidth) {\small agents};
\node[] at (\xB, -0.31\linewidth) {\small time (\si{\second})};
%
% map
\node[anchor=north west] at (0.08\linewidth, 0.055\linewidth)
{\includegraphics[width=0.035\linewidth]{fig/raw/map/#1}};
\node[anchor=north west] at (0.13\linewidth, 0.051\linewidth) {\scriptsize \mapname{#1}};
\node[anchor=north west] at (0.13\linewidth, 0.032\linewidth) {\scriptsize #2 (#3)};
\end{scope}
}
\begin{figure*}[th!]
\centering
\begin{tikzpicture}
\entry{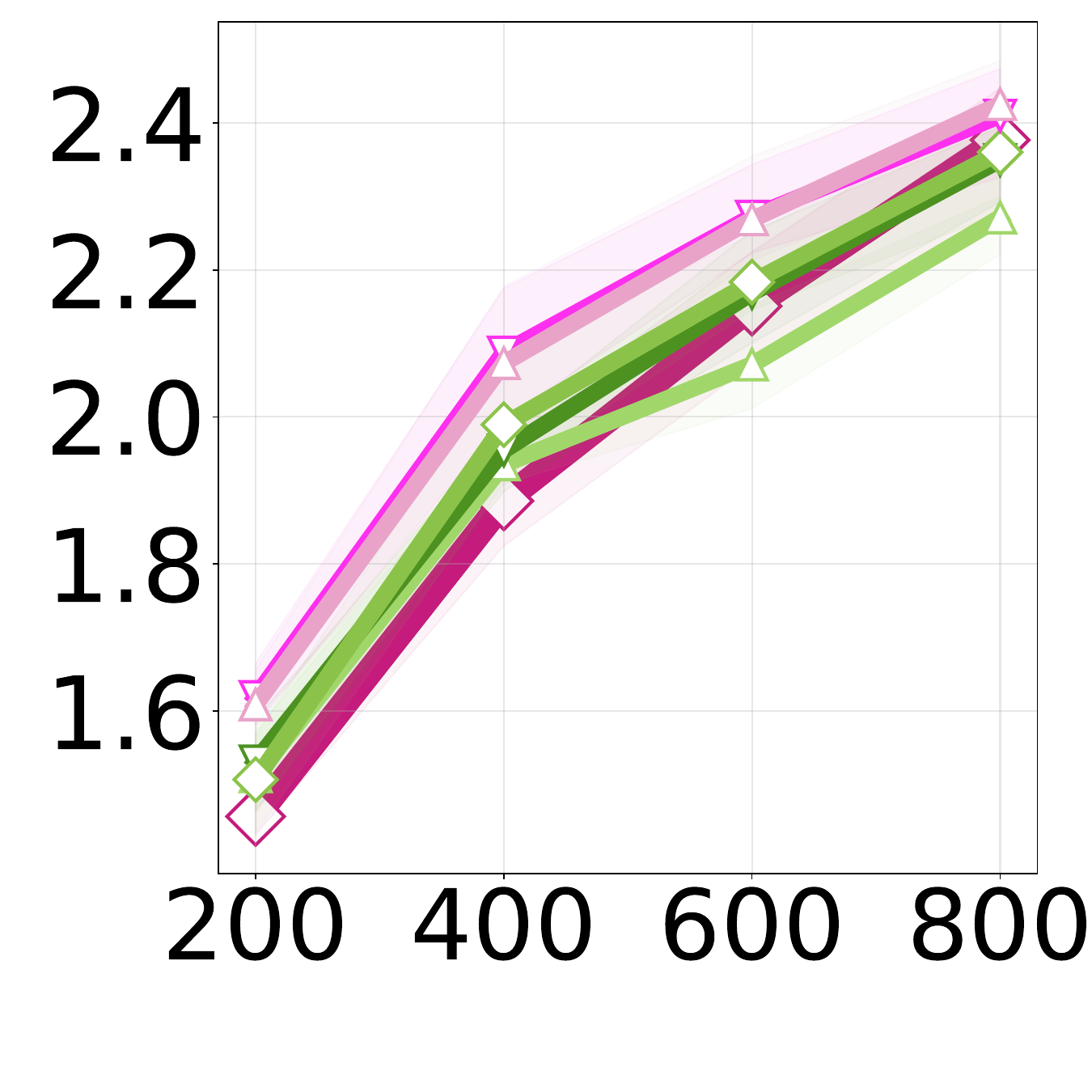}{64x64}{3{,}270}{0}{0}
\entry{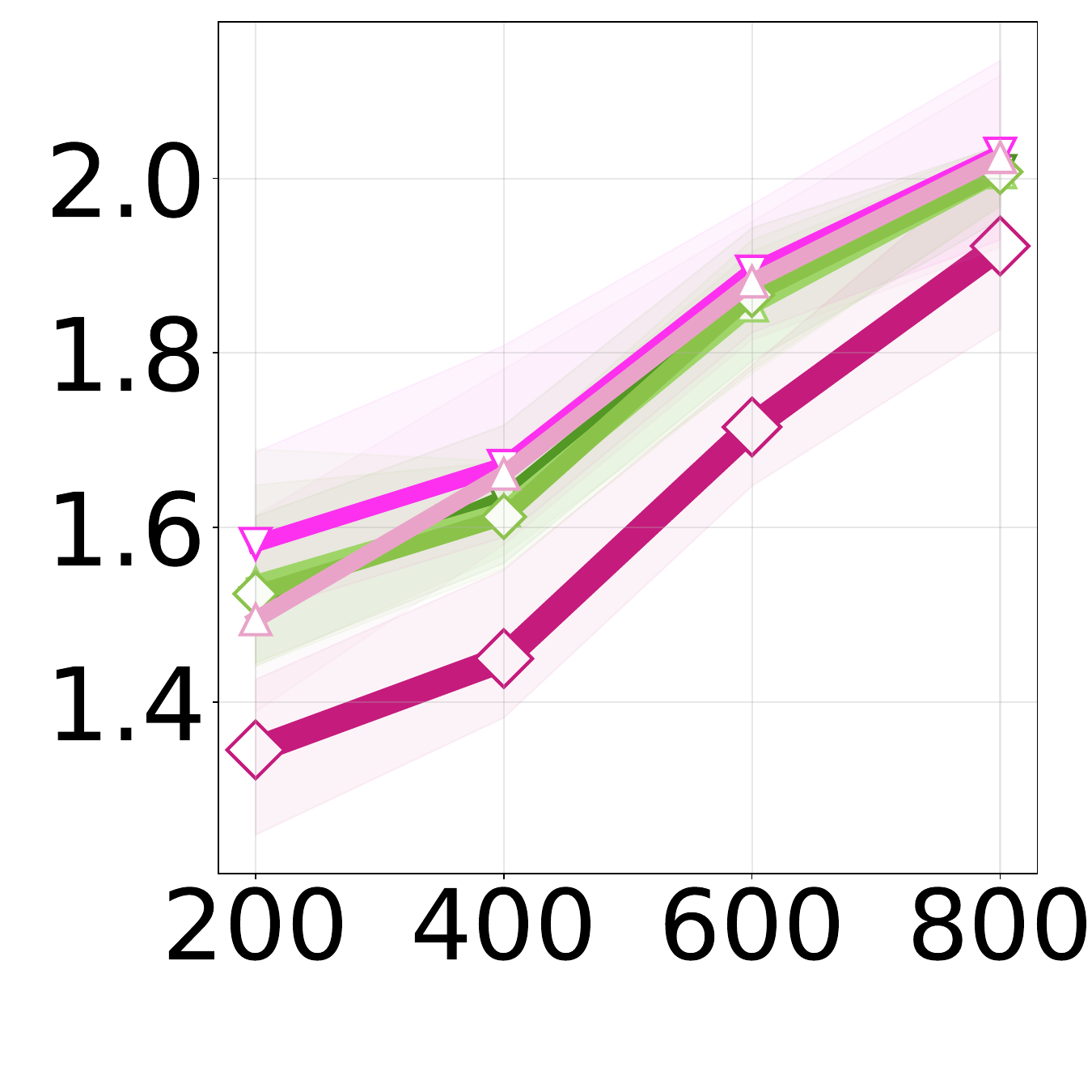}{64x64}{9{,}776}{1}{0}
\entry{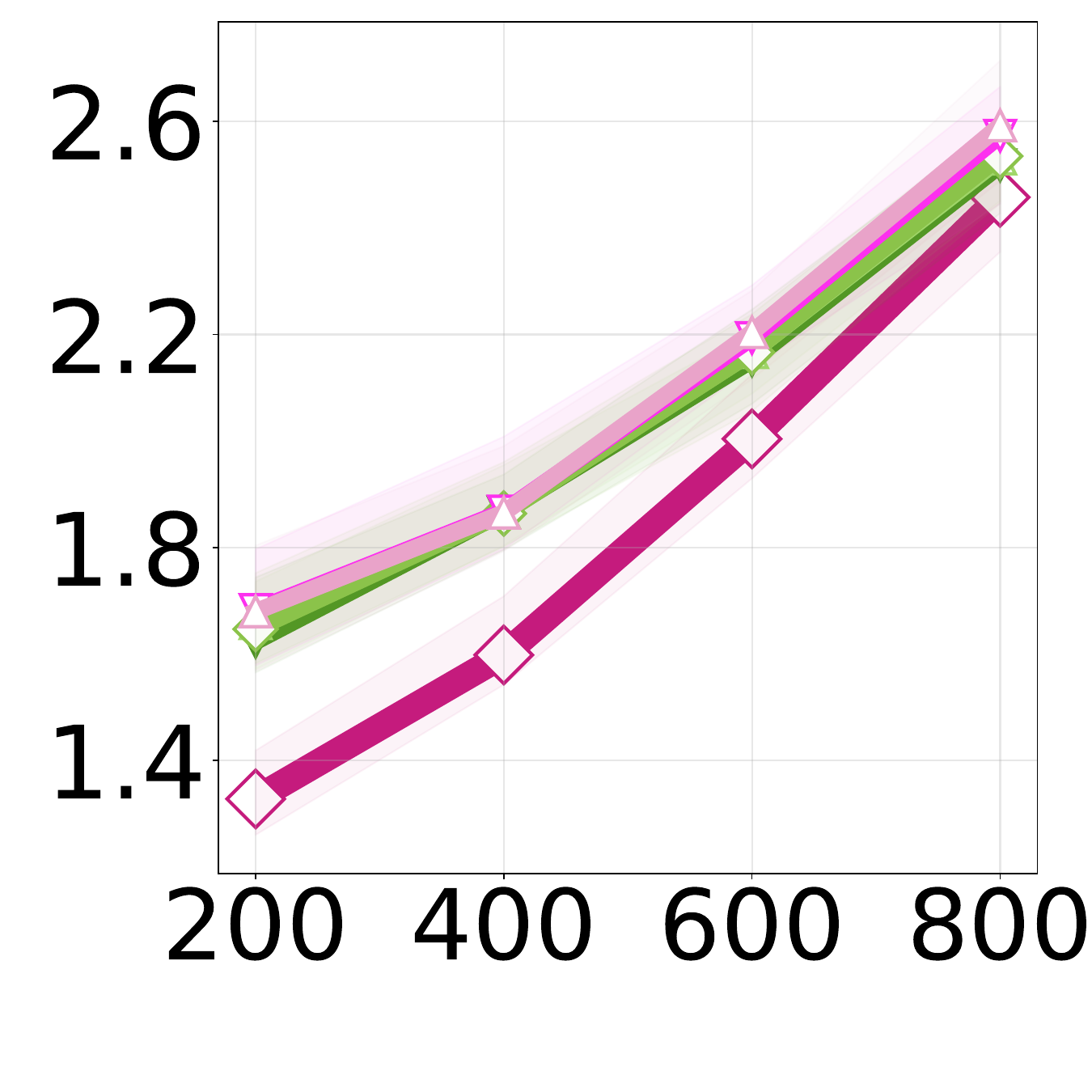}{194x194}{13{,}214}{2}{0}
\entry{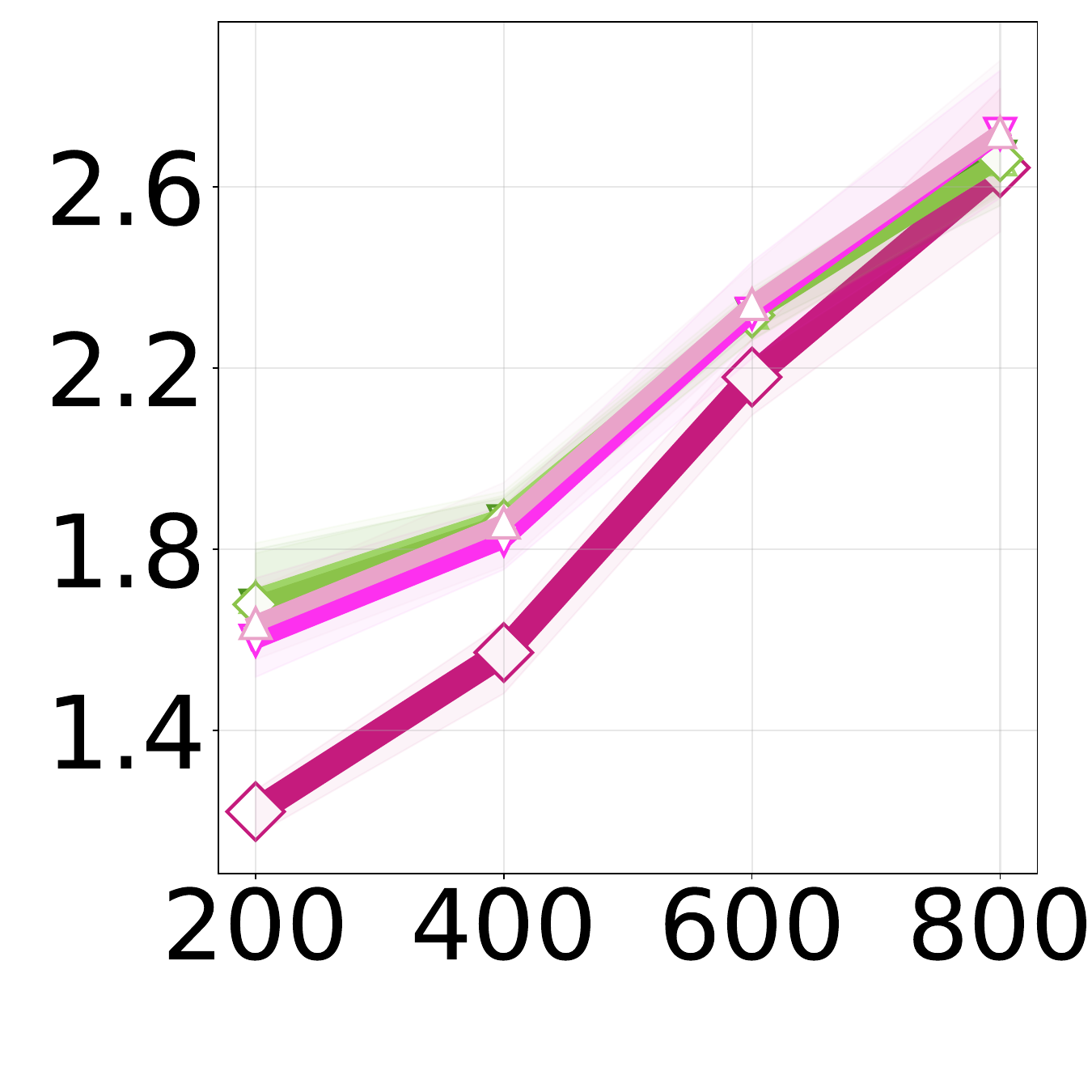}{194x194}{14{,}784}{0}{1}
\entry{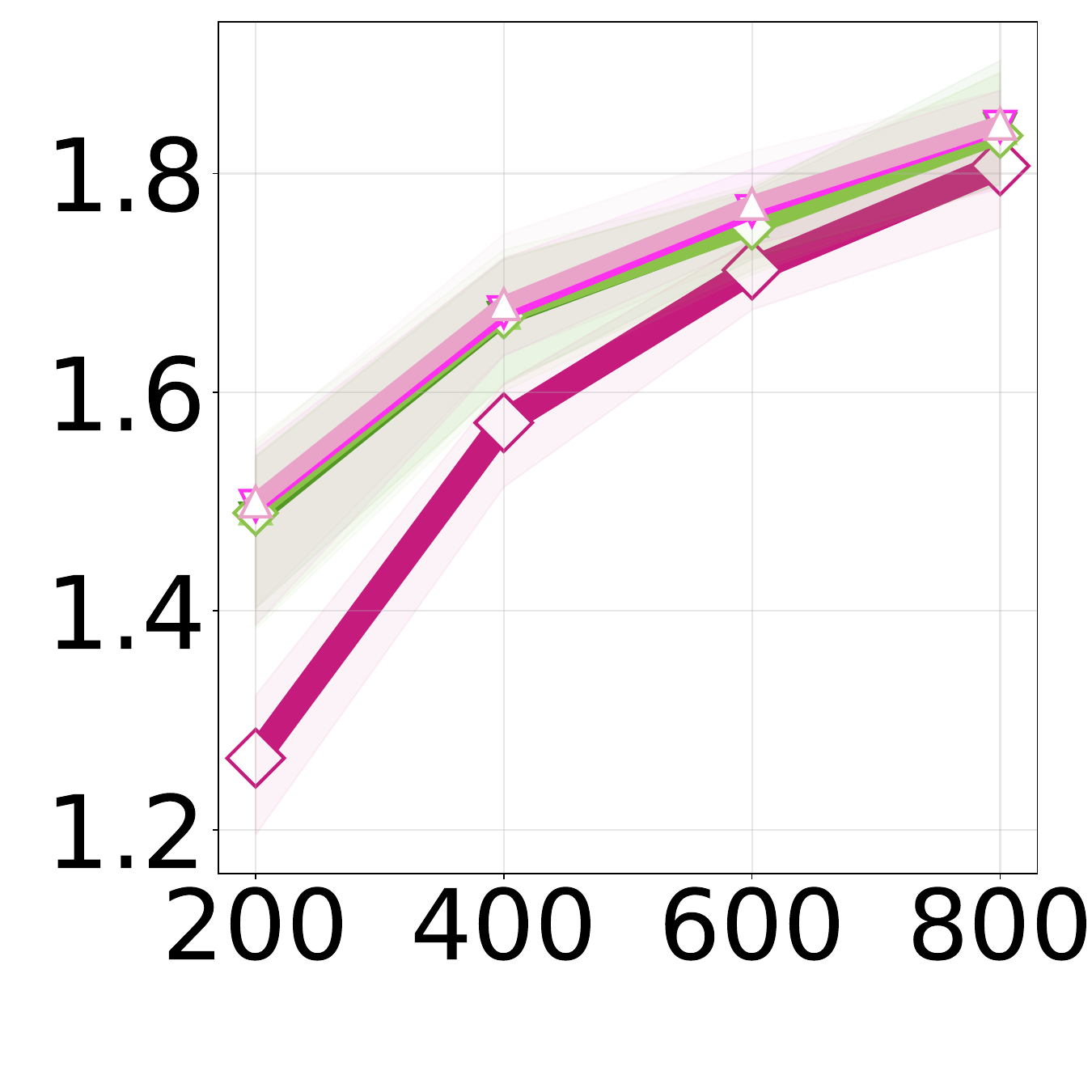}{256x256}{47{,}240}{1}{1}
\entry{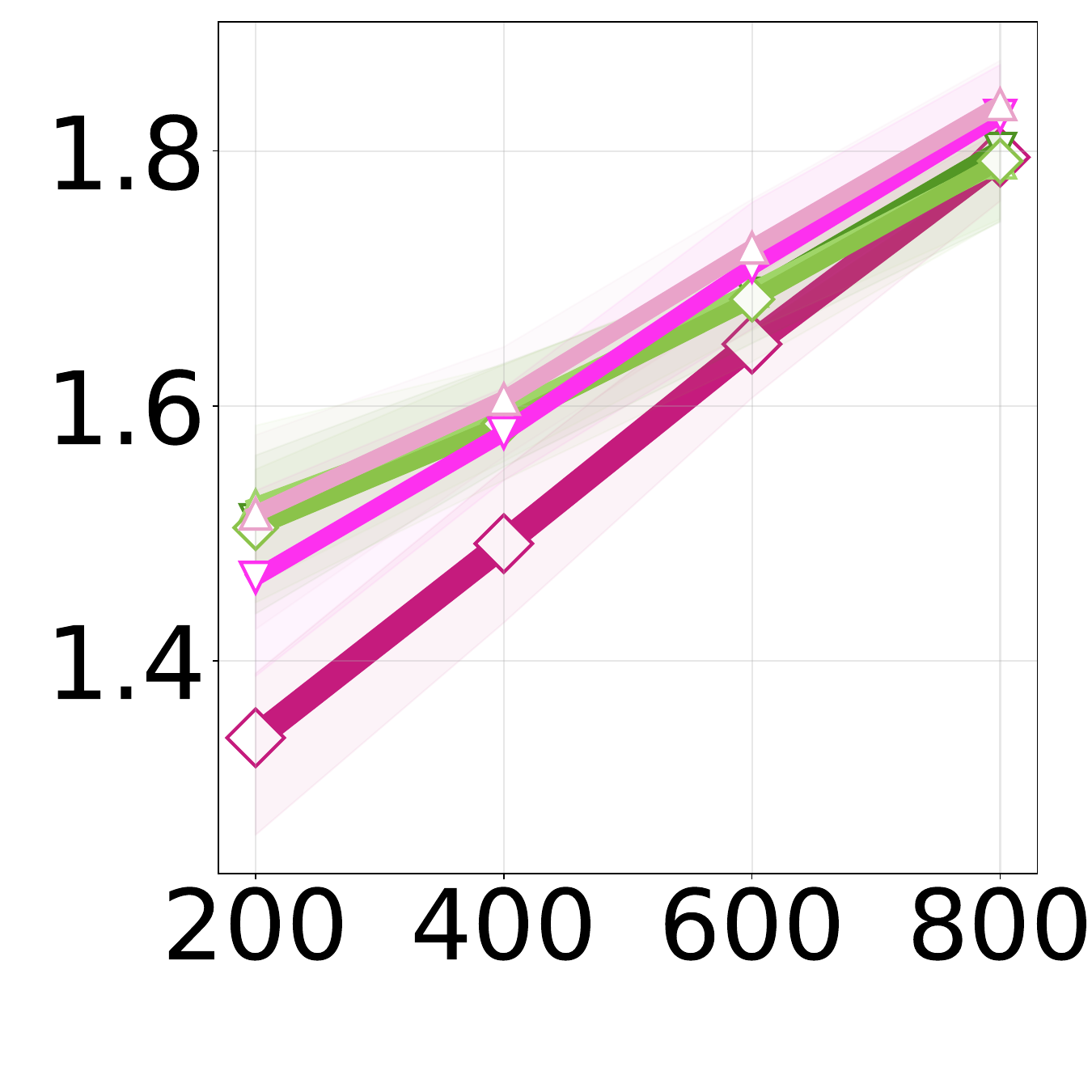}{256x256}{47{,}768}{2}{1}
%
% ylabel
\node[rotate=90] at (-0.1, -0.07\linewidth) {\small ← cost};
\node[rotate=90] at (-0.1, -0.23\linewidth) {\small improvement (\%) →};
\node[rotate=90] at (-0.1, -0.44\linewidth) {\small ← cost};
\node[rotate=90] at (-0.1, -0.61\linewidth) {\small improvement (\%) →};
\end{tikzpicture}
\includegraphics[width=1.0\linewidth]{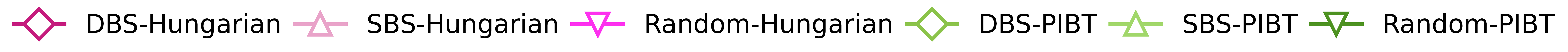}
\caption{
Performance of the proposed TAPF framework using different components.
The number of vertices for each grid is shown in parentheses.
We show the normalized flowtime (median scores), accompanied by minimum and maximum scores using semi-transparent regions.
We further show the improvement rate from the initial solution over time for 200-agent instances.
}
\label{fig:main}
\end{figure*}
}

\section{Evaluation}
\label{sec:evaluation}
This section evaluates the proposed TAPF framework.
The experiments consist of four parts:
\emph{(i)}~identifying effective combinations of reassignment and feedback strategies;
\emph{(ii)}~contrasting our framework against the state-of-the-art TAPF solver;
\emph{(iii)}~assessing its scalability by large instances with up to $10{,}000$ agents; and
\emph{(iv)}~validating the effect of the multi-bottleneck method.

\paragraph{Setup.}
Experiments were conducted on a MacBook Pro with an Apple M1 Chip, equipped with 8-core CPUs and \SI{8}{\giga\byte} RAM.
The proposed methods are coded in Rust.
We benchmark TAPF methods on four-connected grid maps from the MAPF benchmark~\cite{stern2019def}, containing diverse environments with varying obstacle densities and structural characteristics.
Using these maps, we generate TAPF instances with an assignment matrix $\mathcal{T}$, constructed under one of the following two scenarios:
\begin{itemize}
\item \scenRandom: Each agent is assigned ten targets, which are randomly selected from $V$ but placed at a moderate distance from its start position, following the random target placement used in prior TAPF studies~\cite{ITA-ECBS, ita-cbs}.
\item \scenHotspot: Targets for all agents are concentrated in a single densely packed region.
Within this hotspot, each agent is assigned ten targets and creating about $80\%$ target overlap.
\end{itemize}
The former scenario represents sparse, easier-to-solve cases, whereas the latter requires careful target assignment.

By default, the framework uses the multi-bottleneck evaluation method (\cref{algo:multiple_bottleneck}).
For each refinement, it identifies three agents as bottlenecks, whose values were adjusted prior to the experiments to strike a balance between exploration and computational efficiency.
It then computes reassignments and solves the corresponding MAPF with multi-threading.
We employ LaCAM for MAPF solving due to its real-time responsiveness.
As a solution metric, we use the ``cost'' as the flowtime normalized by the sum of the shortest-path lengths between each agent's start and its nearest target, serving as a sub-optimality upper bound (minimum: $1.0$).

\subsection{Framework Design}
The key components of our framework are the feedback mechanism and the target reassignment strategy.
The former can be either DBS or SBS, whereas the latter can use either PIBT or Local Hungarian.
We evaluate six combinations of these components, along with a baseline `Random' strategy that selects an agent uniformly at random, without feedback.
For example, `DBS-Hungarian' denotes the configuration that uses DBS for bottleneck identification and Local Hungarian for target reassignment.
The time limit was set to \SI{10}{\second} and 30 instances were prepared for each setting.
This experiment omits the final path optimization step to focus on the efficacy of different combinations.

\paragraph{Results.}
\Cref{fig:main} summarizes the results.
We first observe that the \scenRandom scenario is easier than \scenHotspot, as spatially distributed targets lead to fewer path intersections (see scales of the y-axes).
Consequently, \scenRandom yields near-optimal initial solutions and leaves only a small margin for improvement.
Nevertheless, DBS-PIBT consistently achieves the best costs in this scenario; even when the overall margin is small, its ability to incrementally improve individual agent assignments accumulates into meaningful gains.
In contrast, \scenHotspot offers more challenging instances where target reassignment has a significant impact.
In either scenario, the initial solutions were obtained rapidly without any failures: e.g., \SI{705}{\milli\second} for 800-agent cases in the worst.
In \scenHotspot scenario, DBS-Hungarian generally achieves the best scores.
For example, in the \scenHotspot results for the 200-agent cases, DBS-Hungarian consistently achieves a 10–25\% improvement over the initial solution, demonstrating superior real-time performance compared to the other methods.
The time profiling of pathfinding and target reassignment is provided in the appendix; regardless of the combination, the pathfinding component dominates the computational cost.

\setlength{\tabcolsep}{3pt}
\begin{table}[t]
  \centering
  \footnotesize
  \begin{tabular}{ll rr rr}
  \toprule
  \multirow{2}{*}{Reassign} & \multirow{2}{*}{Feedback} & \multicolumn{2}{c}{Iter} & \multicolumn{2}{c}{Imprv (\%)} \\
  \cmidrule(lr){3-4} \cmidrule(lr){5-6}
  & & Mean & Min--Max & Mean & Min--Max \\
  \midrule
  & DBS  & 855.3 & 656--980 & \textbf{17.5} & 12.3--21.6 \\
  Hungarian & SBS & 833.4 & 741--887 & 12.2 & 4.7--21.0 \\
  & Random  & 1901.0 & 1654--2032 & 3.6 & 0.0--7.7 \\
  \midrule
  & DBS & 695.1 & 654--730 & 9.4 & 6.2--14.3 \\
  PIBT & SBS & 773.5 & 707--824 & 10.9 & 7.7--13.5 \\
  & Random & 969.1 & 833--1060 & 14.5 & 10.1--21.2 \\
  \bottomrule
  \end{tabular}
  \caption{Framework performance with different components on \mapname{random-64-64-20} with 200 agents within \SI{10}{\second} in \scenHotspot.
    \emph{Iter}: Number of successful target reassignments.
    \emph{Imprv}: Improvement rate from the initial solution.}
  \label{tab:solver_iteration_analysis}
\end{table}
\setlength{\tabcolsep}{3pt}
\begin{table}[t]
  \centering
  \footnotesize
  \begin{tabular}{ll rr rr}
  \toprule
  \multirow{2}{*}{Reassign} & \multirow{2}{*}{Feedback} & \multicolumn{2}{c}{Time (\si{\milli\second})} & \multicolumn{2}{c}{Imprv (\%)} \\
  \cmidrule(lr){3-4} \cmidrule(lr){5-6}
  & & Mean & Min--Max & Mean & Min--Max \\
  \midrule
  & DBS & 2026 & 1847--2297 & \textbf{17.4} & 10.9--22.7 \\
  Hungarian & SBS & 2125 & 1943--2428 & 12.2 & 0.0--15.9 \\
  & Random  & 934 & 883--1044 & 4.8 & 0.0--9.8 \\
  \midrule
  & DBS & 2839 & 2705--3203 & 8.6 & 4.0--14.3 \\
  PIBT & SBS & 2754 & 2517--3249 & 7.5 & 5.2--12.5 \\
  & Random & 2016 & 1803--2466 & 10.7 & 6.3--19.1 \\
  \bottomrule
  \end{tabular}
  \caption{
Fixing the number of iterations to 100, using the same setting of \cref{tab:solver_iteration_analysis}.
`Time' taken to reach the iteration limit is shown.
}
  \label{tab:max-iter-analysis}
\end{table}

{
\setlength{\tabcolsep}{3pt}
\begin{table}[t]
  \centering
  \footnotesize
  \begin{tabular}{llc cr rr}
  \toprule
  \multirow{2}{*}{Reassign} & \multirow{2}{*}{Feedback} & \multirow{2}{*}{Cost Gap}
  & Iter & \multicolumn{2}{c}{Imprv (\%)} \\
  \cmidrule(lr){4-4} \cmidrule(lr){5-6}
  & & & Mean  & Mean & Min--Max \\
  \midrule
  & DBS & 1.26 & 447.0 & 7.4 & 4.5--14.1 \\
  Hungarian & SBS & 1.03 & 463.3 & \textbf{19.0} & 12.7--26.3 \\
  & Random & 1.06 & 1081.3 & 8.5 & 1.8--13.2 \\
  \midrule
  & DBS & 1.09 & 351.6 & 8.7 & 4.6--10.8 \\
  PIBT & SBS & 1.10 & 375.4 & 9.4 & 5.1--12.9 \\
  & Random & 1.06 & 518.9 & 15.5 & 11.4--21.9 \\
  \bottomrule
  \end{tabular}
  \caption{Framework performance using random initial assignment.
  The setting follows \cref{tab:solver_iteration_analysis}.
  `Cost Gap' shows the final solution costs, divided by those obtained with the greedy initial assignment (\cref{tab:solver_iteration_analysis}).
  All values exceed 1.0, meaning that no algorithm achieves the same quality under random initial assignment.
  }
  \label{tab:random_initial_analysis}
\end{table}
}

For further investigation, \cref{tab:solver_iteration_analysis,tab:max-iter-analysis} provide detailed statistics for the six combinations on \mapname{random-64-64-20} with 200 agents in \scenHotspot.
\Cref{tab:solver_iteration_analysis} fixes the time budget, whereas \cref{tab:max-iter-analysis} fixes the number of reassignment iterations.
Focusing on PIBT, Random-PIBT achieves the highest number of iterations (\cref{tab:solver_iteration_analysis}) and the fastest per-iteration speed (\cref{tab:max-iter-analysis}), while also delivering the best improvement rate in both settings. This indicates that, for PIBT, exploring a larger number of diverse target combinations matters more than leveraging feedback-guided agent selection.
\Cref{tab:max-iter-analysis} further supports that even when the number of iterations is fixed, Random-PIBT yields the highest improvement rate with the least computation time.
Turning to Hungarian-based methods, they achieve a better solution quality.
Random-Hungarian achieves the highest number of iterations due to its simplicity, while its improvement remains lower than the other two.
Meanwhile, both DBS-Hungarian and SBS-Hungarian achieve low iteration counts but deliver substantial improvements, although DBS generally outperforms SBS in both per-iteration speed and improvement efficacy.
This is somewhat surprising, given the expectation that SBS would capture more nuanced bottleneck agents than DBS.
Based on these observations, we use DBS-Hungarian in the following experiments.

\paragraph{Sensitivity to Initial Target Assignment.}
Before moving on, we investigate how the initial target assignment (\cref{algo:framework:initial-assignment} in \cref{algo:framework}) affects the outcomes.
Using the same experimental setting as in \cref{tab:solver_iteration_analysis}, \cref{tab:random_initial_analysis} presents the empirical results obtained when a completely random initial target assignment is used.
The results demonstrate that the choice of initial assignment has a substantial impact on the final solution quality: all Cost Gap values exceed 1.0, indicating that the random initial assignment consistently yields inferior final solutions compared to the greedy initial assignment, despite the subsequent iterative improvement.
The relative performance of the framework’s components differs from that in \cref{tab:solver_iteration_analysis}, where SBS-Hungarian achieves the greatest improvements.
We conjecture that the random assignment makes the corresponding MAPF problem harder by introducing many path intersections among the agents’ shortest paths. As a result, SBS is able to identify more effective bottleneck sets than DBS.
Conversely, the strong performance of DBS-Hungarian observed so far also depends on having an effective initial assignment.

\subsection{Final Path Optimization}
\begin{figure}[t!]
    \raggedright
    \begin{minipage}[c]{1\columnwidth}
        \centering
        \includegraphics[width=\linewidth]{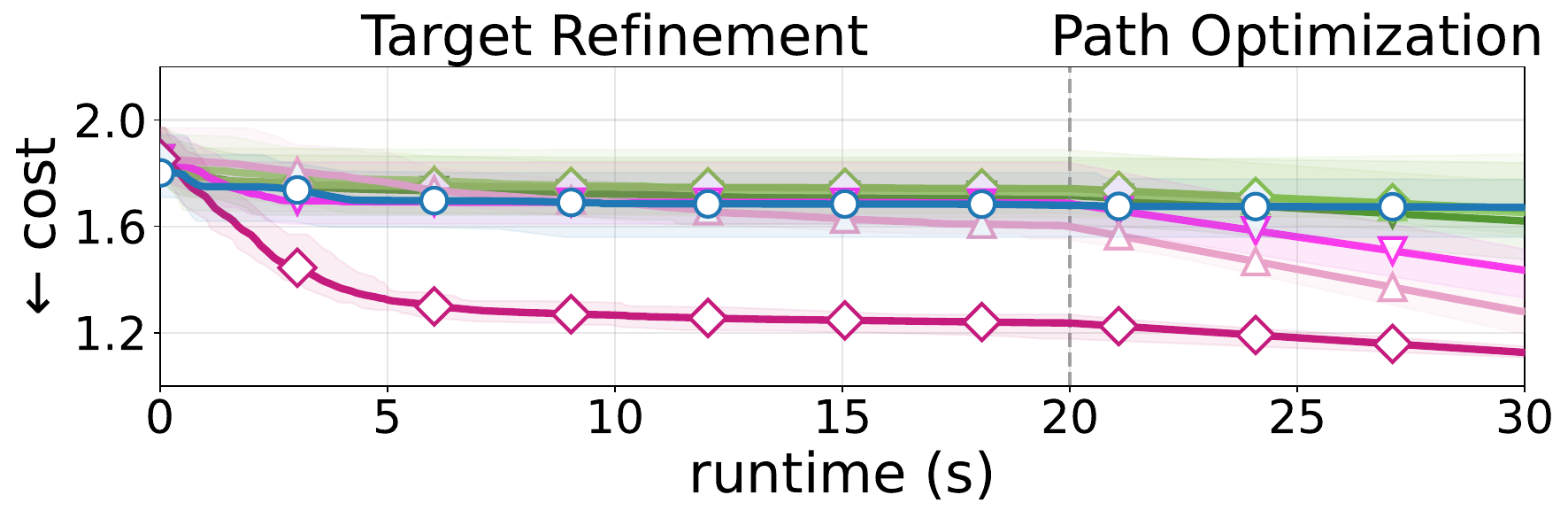}
    \end{minipage}
 \centerline{\includegraphics[width=1.0\linewidth]{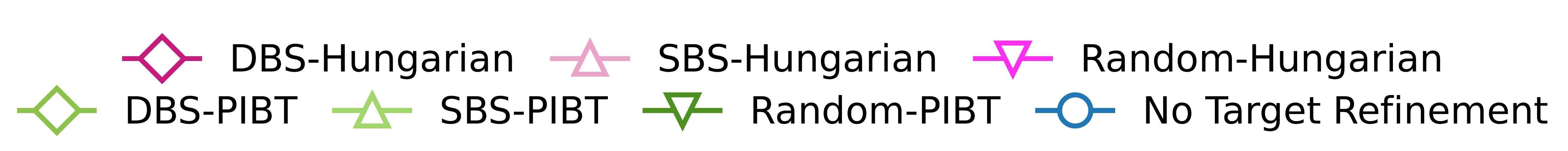}}
    \caption{Framework performance on \mapname{lak303d} with 200 agents in \scenHotspot, using \SI{20}{\second} target refinement followed by \SI{10}{\second} final path optimization.
  \emph{No Target Refinement} skips target reassignments and allocates the full time for path optimization.
  }
    \label{fig:optimization}
\end{figure}

The preceding experiments omitted the path optimization step to isolate the efficacy of each component combination.
\Cref{fig:optimization} shows that the path optimization consistently improves solution quality across all method combinations, though the magnitude varies sharply: Hungarian-based methods benefit substantially, while PIBT-based methods see only modest gains.
\emph{No Target Refinement}, which allocates the full time budget to LaCAM3's anytime path optimization, also improves the initial solution but plateaus above every combination that includes target refinement.
This confirms that solving TAPF efficiently requires the interplay of target refinement and path optimization, not path planning alone.

\subsection{Comparison with ITA-ECBS}
We compare DBS-Hungarian with ITA-ECBS~\cite{ITA-ECBS}, a leading TAPF solver.
Both methods are given a time limit of \SI{30}{\second}.
For DBS-Hungarian, we allocated \SI{20}{\second} for iterative refinement and \SI{10}{\second} for the final path optimization using LaCAM3's anytime search~\cite{okumura2024lacam3}.

\Cref{table:comp-itaecbs} highlights the scalability advantage of DBS-Hungarian.
DBS-Hungarian achieves a 100\% success rate across all settings, whereas ITA-ECBS fails on the majority of instances once the number of agents exceeds 20 in \scenHotspot and roughly 100 in \scenRandom, as the CBS conflict tree grows exponentially with agent count.
As shown in \cref{fig:main}, DBS-Hungarian scales to 800 agents in \scenHotspot, far beyond the reach of ITA-ECBS.
Moreover, while DBS-Hungarian yields slightly higher costs than ITA-ECBS, the gap is small, indicating competitive solution quality.

{
\setlength{\tabcolsep}{3pt}
\newcommand{\cmid}{\cmidrule(ll){1-7}}

\begin{table}[t!]
    \footnotesize
    \centering
    \begin{tabular}{llrrrrrrrrrr}
      \toprule
      &
      & & \multicolumn{2}{c}{success rate(\%)}
      & \multicolumn{2}{c}{cost}
      \\ \cmidrule(lr){4-5}\cmidrule(lr){6-7}
      map & scenario & $|A|$
      & ECBS & D-H
      & ECBS & D-H \\ \midrule
      \multirow{3}{*}{\makecell[l]{\mapname{random-}\\[-0.5ex]\mapname{32-32-20}}}
      & \multirow{2}{*}{\scenHotspot} & 10 & 100 & 100 & 1.015 & 1.055 \\
      && 20 & 20 & 100 & 1.033 & 1.045 \\ \cmidrule(r){2-7}
      & \scenRandom & 50  & 50 & 100 & 1.003 & 1.029 \\ \midrule
      \multirow{5}{*}{\mapname{den312d}}
      & \multirow{2}{*}{\scenHotspot} & 10 & 70 & 100 & 1.004 & 1.027 \\ 
      && 20 & 30 & 100 & 1.011 & 1.089 \\ \cmidrule(r){2-7}
      & \multirow{3}{*}{\scenRandom}
      & 50  & 93 & 100 & 1.001 & 1.005 \\
      && 100 & 40 & 100 & 1.001 & 1.011 \\
      && 150 & 13 & 100 & 1.001 & 1.018 \\ \midrule
     \multirow{7}{*}{\makecell[l]{\textit{warehouse-}\\[-0.5ex]\mapname{10-20-10-2-1}}}
      & \multirow{3}{*}{\scenHotspot} & 10 & 100 & 100 & 1.003 & 1.029 \\
      && 20 & 90 & 100 & 1.002 & 1.036 \\
      && 30 & 60 & 100 & 1.003 & 1.043 \\ \cmidrule(r){2-7}
      & \multirow{4}{*}{\scenRandom}
      & 50  & 100 & 100 & 1.001 & 1.005 \\
      && 100  & 87 & 100 & 1.002 & 1.007 \\
      && 150 & 13 & 100 & 1.001 & 1.014 \\
      && 200 & 3 & 100 & 1.002 & 1.019 \\ 
      \bottomrule
    \end{tabular}
\caption{
Comparison of ITA-ECBS (ECBS) and our proposed DBS-Hungarian (D-H).
All scores are averages over solved instances by both.
For instance preparation, we gradually increased the number of agents until ITA-ECBS failed to solve all instances among 30 instances generated.
}
\label{table:comp-itaecbs}
\end{table}
}

\subsection{Scalability Test}
\begin{figure}[t!]
    \raggedright
    \begin{minipage}[c]{1\columnwidth}
        \centering
        \includegraphics[width=\linewidth]{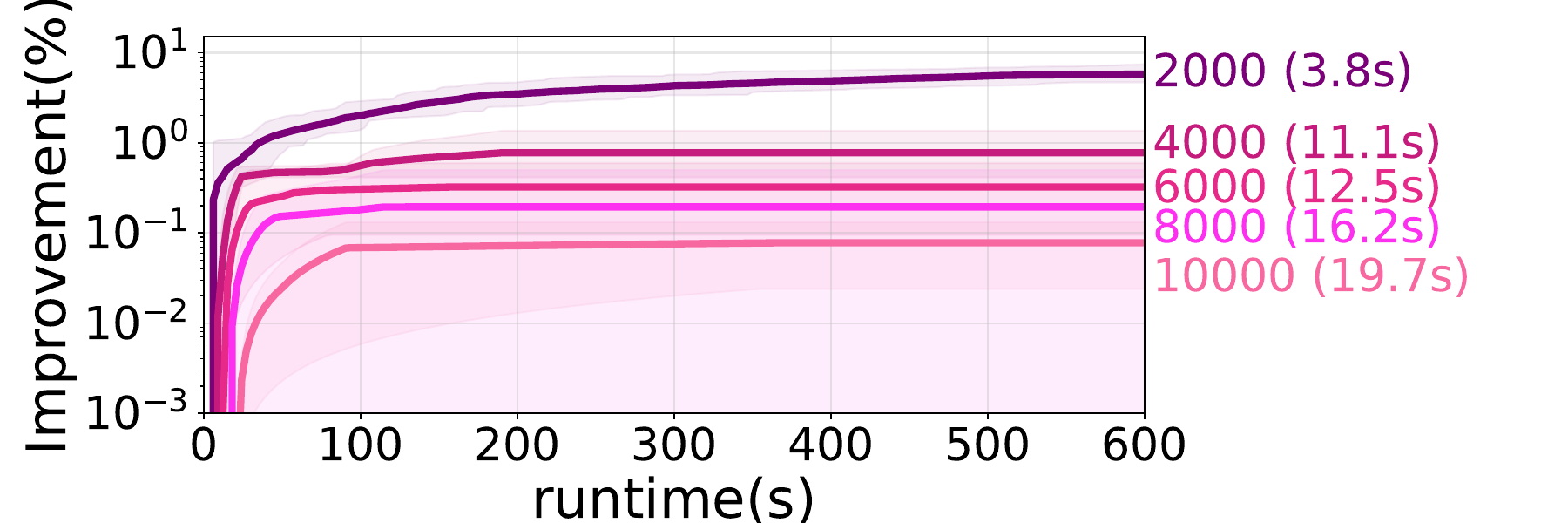}
    \end{minipage}
    \caption{Scalability test.
    On the right, \#agents and the time for finding the initial solutions (in parentheses) are shown.}
    \label{fig:scalability}
\end{figure}

We evaluate our framework with up to $10{,}000$ agents, using \mapname{warehouse-20-40-10-2-2}.
The time limit was set to \SI{10}{\minute} given the problem scales.
\Cref{fig:scalability} shows the time for finding initial solutions and the improvement rate over time.
Our method finds initial solutions around \SI{20}{\second} even for $10{,}000$ agents, and continues the solution refinement.
Although the improvement becomes more modest as the number of agents increases, the framework consistently reduces the cost, demonstrating its ability to handle massive-scale TAPF instances while continuing to refine solutions.

\subsection{Effect of Multi-Bottleneck Selection}
\label{sec:ablation}
Finally, we investigate the effect of the number of bottleneck agents evaluated per iteration, herein denoted as $k \in \{1, 3, 10\}$.
$k=1$ corresponds to the vanilla framework (\cref{algo:framework}), while the other corresponds to \cref{algo:multiple_bottleneck}.
We fix the bottleneck identification method to DBS and assess the effect of $k$ using both the PIBT and Local Hungarian.

{
\setlength{\tabcolsep}{0.5pt}
\newcommand{\entry}[2]{
\begin{minipage}{0.23\linewidth}
\centering
\begin{tikzpicture}
\node[anchor=north] at (0, 0)
{\includegraphics[width=1.0\linewidth]
{fig/raw/ablation/#1_#2_ost003d.pdf}};
\end{tikzpicture}
\end{minipage}
}
\begin{figure}[t!]
  \centering
  \begin{tabular}{cccc}
    \multicolumn{2}{c}{DBS-PIBT}&
    \multicolumn{2}{c}{DBS-Hungarian}\\
    \entry{pibt}{imprv} &
    \entry{pibt}{iter} &
    \entry{hungarian}{imprv} &
    \entry{hungarian}{iter}
    \\
    \quad~~agents & \multicolumn{3}{c}{\quad\includegraphics[width=0.55\linewidth,clip,trim={0 1.25cm 0 0.9cm}]{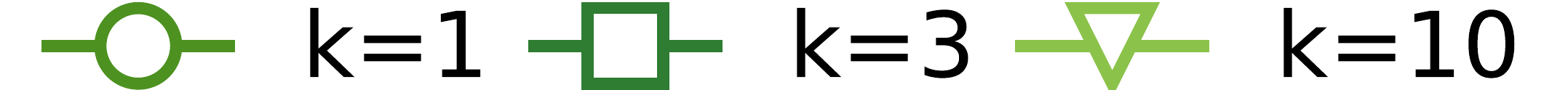}}
  \end{tabular}
  \caption{Effect of the number of agents for target reassignment ($k$) on DBS-\{PIBT, Hungarian\} using \mapname{ost003d}.
    Improvement rates from the initial solutions and the number of iterations completed within \SI{10}{\second} are shown.
    }
\label{fig:ablation_k}
\end{figure}
}

\paragraph{PIBT.}
\Cref{fig:ablation_k} (left) shows that $k=3$ achieves the best improvement across all scenarios.
$k=10$ completes significantly fewer iterations due to computational overhead, resulting in smaller improvements.
$k=1$ performs comparably to $k=3$ up to 200 agents but degrades beyond that due to insufficient exploration diversity.

\paragraph{Local Hungarian.}
\Cref{fig:ablation_k} (right) again shows that $k=3$ as the best choice.
$k=10$ suffers from high per-iteration cost, limiting the total number of iterations, while $k=1$ enables many iterations but each reassignment is too narrow to yield substantial gains.

\section{Conclusion}
We introduced an iterative refinement framework for TAPF that decouples target assignment and pathfinding, achieving scalability advantages over the leading TAPF solver while maintaining competitive solution quality.
Within this framework, we presented two target reassignment methods (PIBT and Local Hungarian) and two feedback mechanisms (DBS and SBS).
Our experiments show that combining DBS and Local Hungarian generally the most effective.

For future work, an interesting direction is to adaptively select the mechanism and reassignment strategy during the refinement loop, akin to the adaptive operator selection in LNS~\cite{li2021anytime}.
\section*{Acknowledgments}
This research was partially supported by JSPS KAKENHI Grant Number 25K21289, JST ACT-X (JPMJAX22A1), and JST PRESTO (JPMJPR2513).
\bibliography{sty/ref-macro,ref}
\newpage
\appendix
\onecolumn
\section*{Appendix}

\section{Time profiling of pathfinding and reassignment}
\begin{figure*}[hp!]
    \centering
    \raisebox{2em}{\rotatebox{90}{\small Runtime (ms)}}\hspace{0.5em}%
    \begin{subfigure}[b]{0.15\linewidth}
        \centering
        \small \quad DBS-Hungarian
        \includegraphics[width=\linewidth]{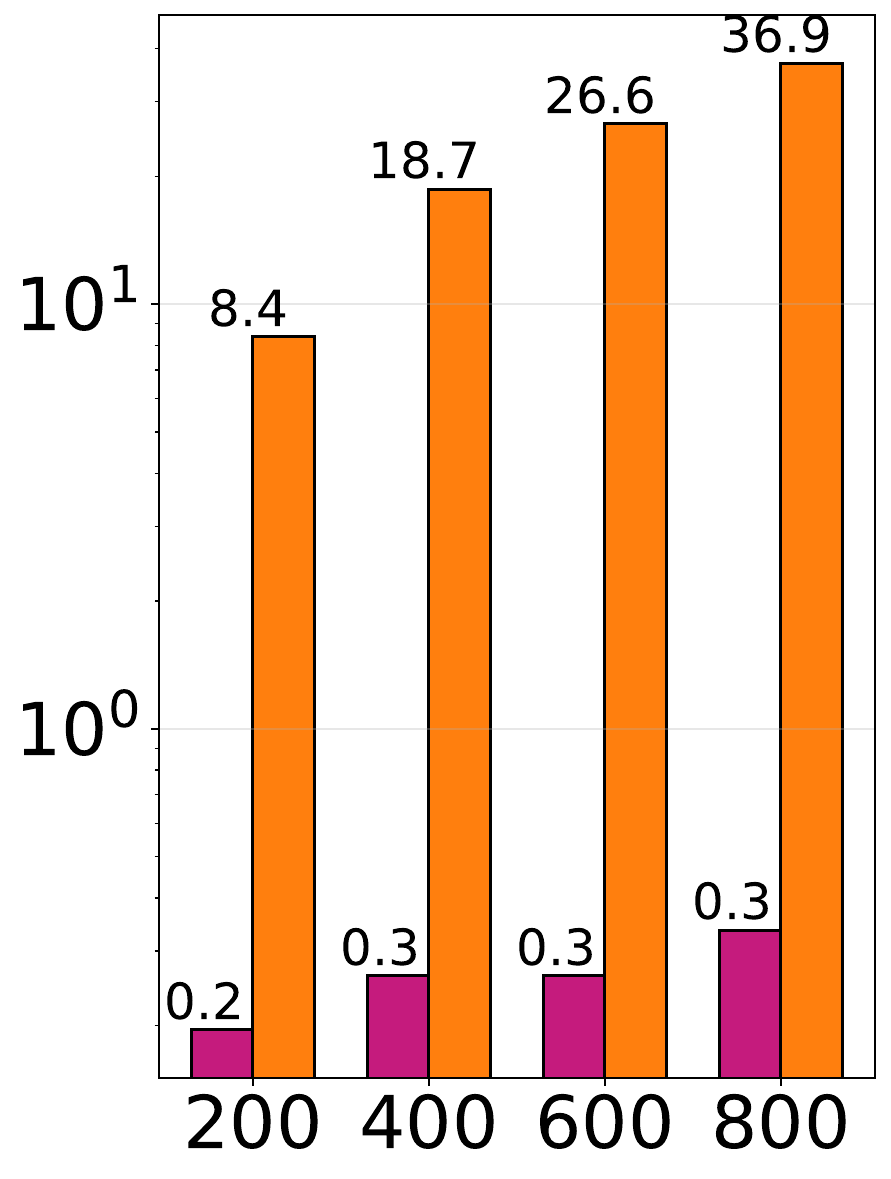}
    \end{subfigure}
    \hfill
    \begin{subfigure}[b]{0.15\linewidth}
        \centering
        \small \quad SBS-Hungarian
        \includegraphics[width=\linewidth]{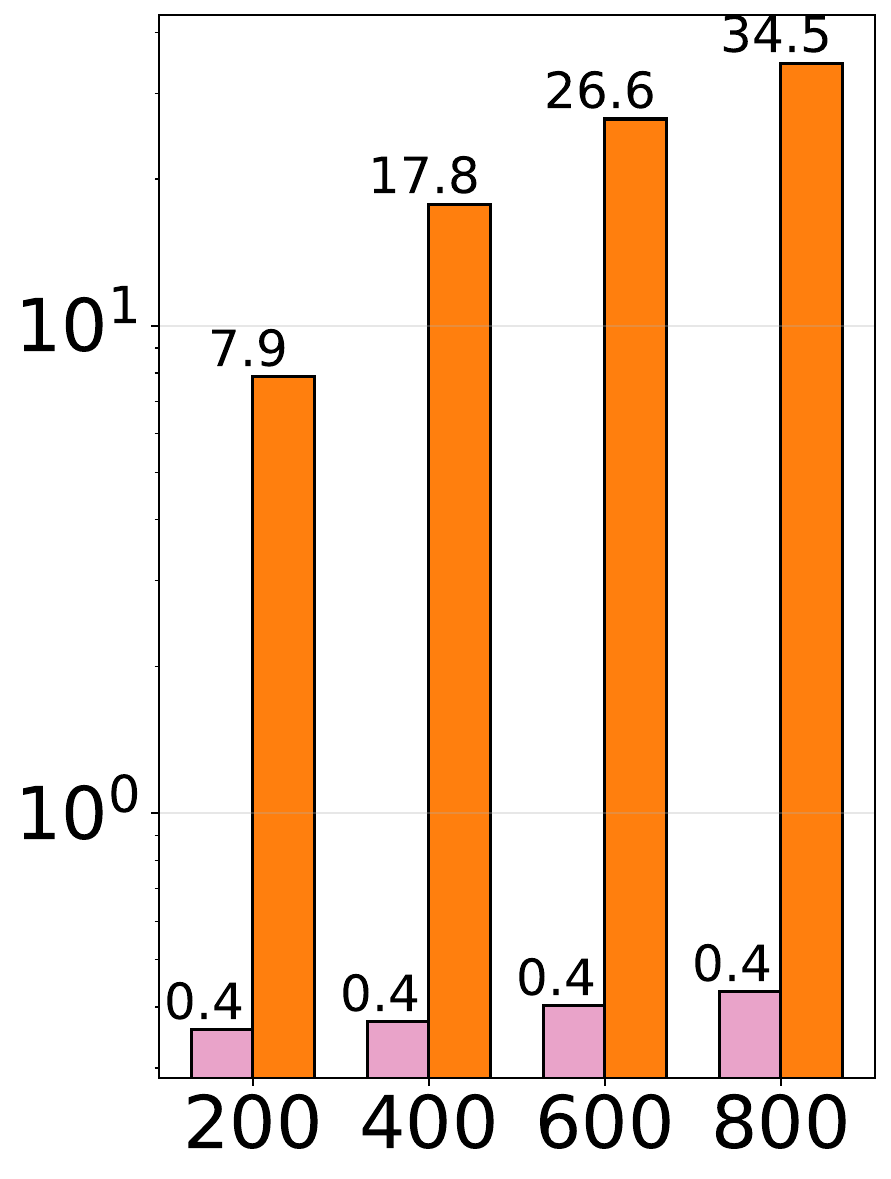}
    \end{subfigure}
    \hfill
    \begin{subfigure}[b]{0.15\linewidth}
        \centering
        \small \quad Random-Hungarian
        \includegraphics[width=\linewidth]{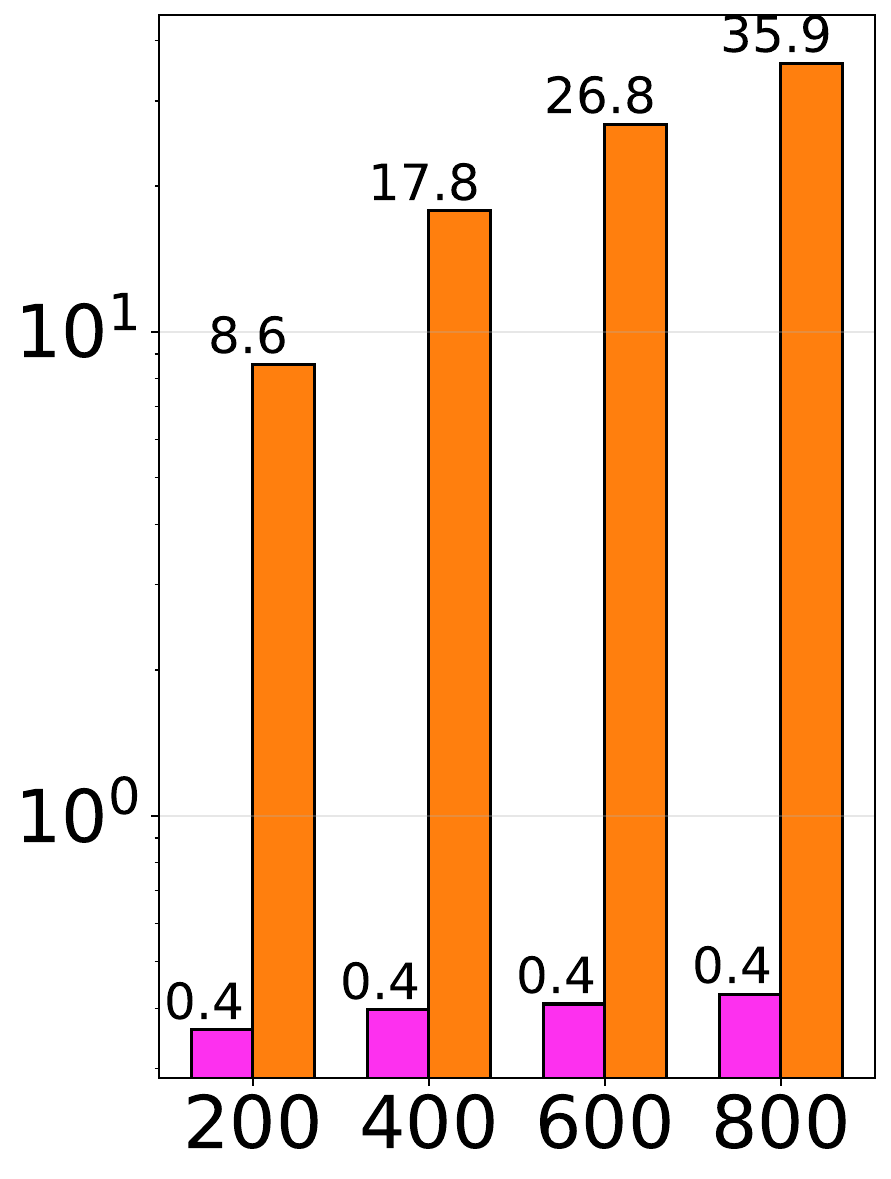}
    \end{subfigure}
    \hfill
    \begin{subfigure}[b]{0.15\linewidth}
        \centering
        \small \quad DBS-PIBT
        \includegraphics[width=\linewidth]{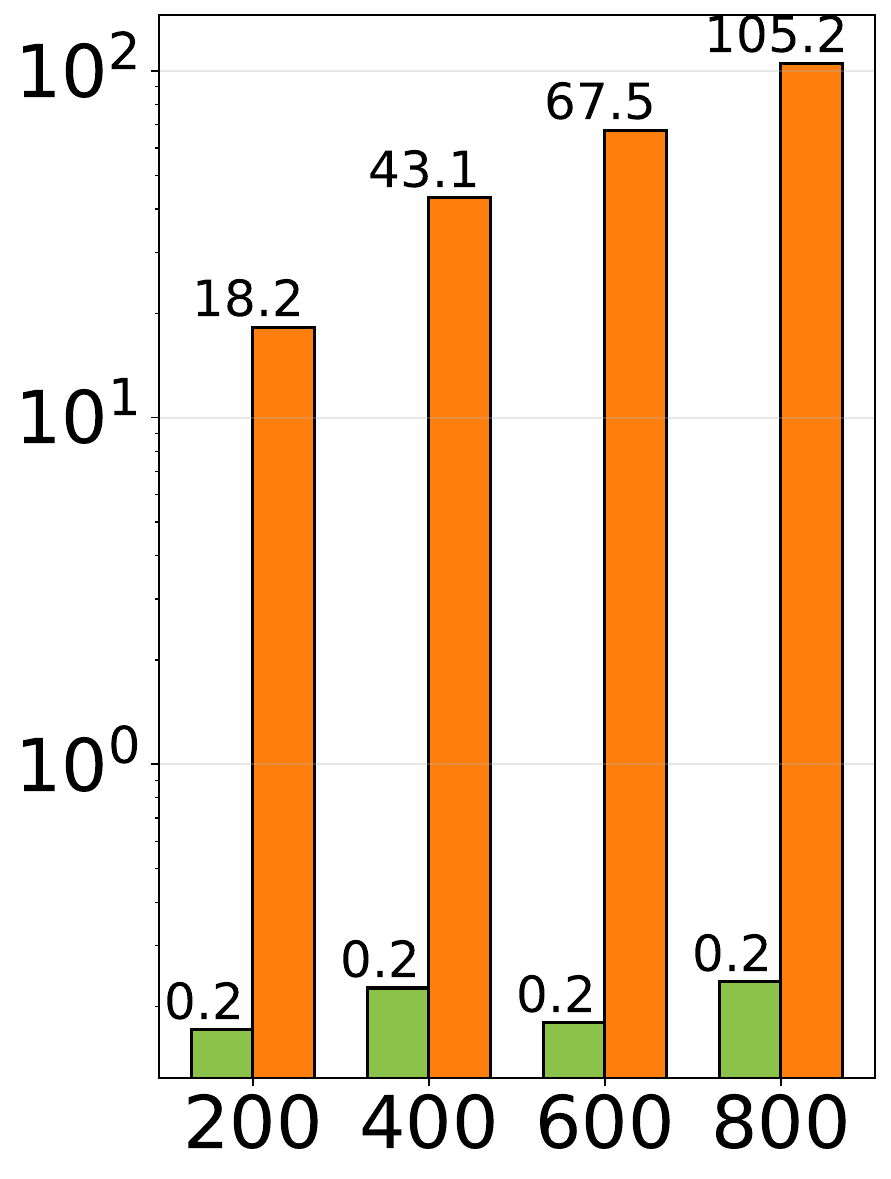}
    \end{subfigure}
    \hfill
    \begin{subfigure}[b]{0.15\linewidth}
        \centering
        \small \quad SBS-PIBT
        \includegraphics[width=\linewidth]{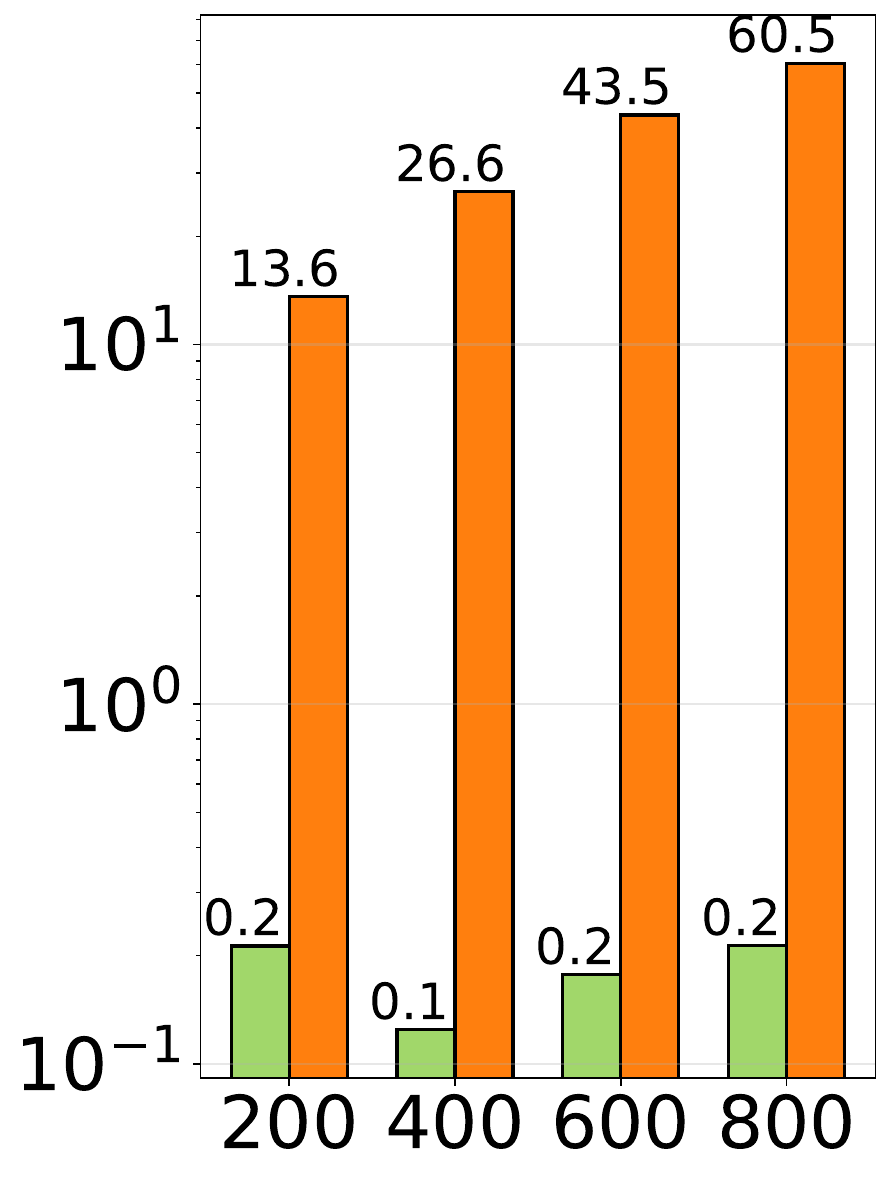}
    \end{subfigure}
    \hfill
    \begin{subfigure}[b]{0.15\linewidth}
        \centering
        \small \quad Random-PIBT
        \includegraphics[width=\linewidth]{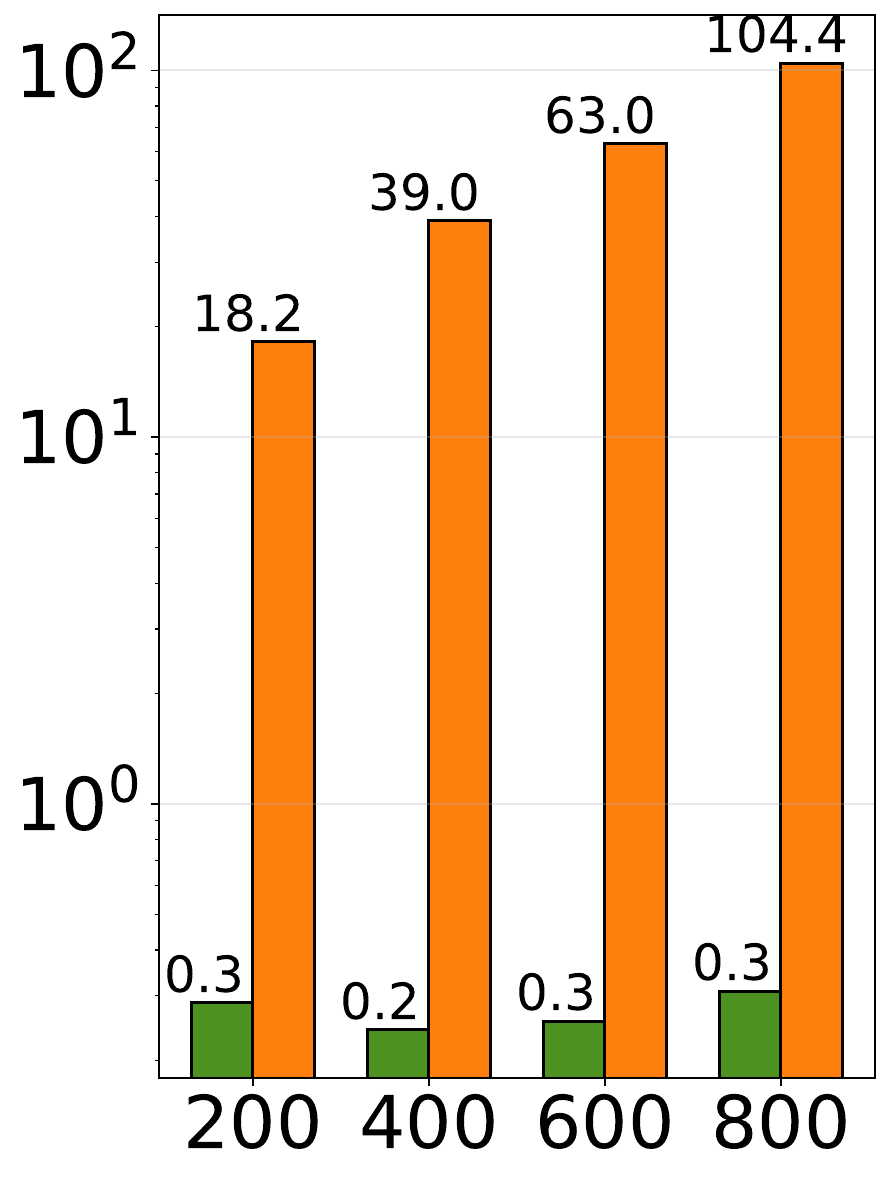}
    \end{subfigure}
    \centering
    agents
    \caption{Time comparison between pathfinding and reassignment on map \mapname{random-64-64-20} in \scenHotspot.
    The left bars show the time spent on pathfinding during iterations, while the right orange bars (\textcolor{orange}{$\blacksquare$}) show the time on pathfinding.
    }
    \label{fig:lacam_reassignment}
\end{figure*}

\Cref{fig:lacam_reassignment} shows that the time spent on pathfinding increases gradually as the number of agents grows, while the time for reassignment remains constant.
Although LaCAM quickly finds the paths, the pathfinding effort is non-negligible.

\end{document}